\pdfoutput=1

\documentclass[11pt]{article}

\usepackage[]{acl}

\usepackage{times}
\usepackage{latexsym}
\usepackage{bm}
\usepackage{multirow}
\usepackage{graphicx}
\usepackage{amsmath}
\usepackage{amssymb}
\usepackage{subfigure}
\usepackage{graphicx} 
\usepackage{float}
\usepackage{stfloats}

\usepackage[T1]{fontenc}

\usepackage[utf8]{inputenc}

\usepackage{microtype}

\title{Supporting Medical Relation Extraction via Causality-Pruned Semantic Dependency Forest}

\author{ Yifan Jin$^{1,2,}$\thanks{\ \ 
Equal contribution.}, Jiangmeng Li$^{1,2,*,}$\thanks{\ \  Corresponding author.}, Zheng Lian$^{1,2}$, Chengbo Jiao$^3$, Xiaohui Hu$^{1,2}$
 \\$^1$University of Chinese Academy of Sciences\\
 $^2$ Institute of Software Chinese Academy of Sciences\\
 $^3$University Of Electronic Science And Technology Of China
 \\\{yifan2020,jiangmeng2019,lianzheng2017,hxh\}@iscas.ac.cn
 \\chengbojiao@hotmail.com}

\begin{document}
\maketitle
\begin{abstract}
Medical Relation Extraction (MRE) task aims to extract relations between entities in medical texts. Traditional relation extraction methods achieve impressive success by exploring the syntactic information, e.g., dependency tree. However, the quality of the 1-best dependency tree for medical texts produced by an out-of-domain parser is relatively limited so that the performance of medical relation extraction method may degenerate. To this end, we propose a method to jointly model semantic and syntactic information from medical texts based on causal explanation theory. We generate dependency forests consisting of the semantic-embedded 1-best dependency tree. Then, a task-specific causal explainer is adopted to prune the dependency forests, which are further fed into a designed graph convolutional network to learn the corresponding representation for downstream task. Empirically, the various comparisons on benchmark medical datasets demonstrate the effectiveness of our model.
\end{abstract}

\section{Introduction}
\begin{figure*}[t]
\centering
\includegraphics[scale=0.25]{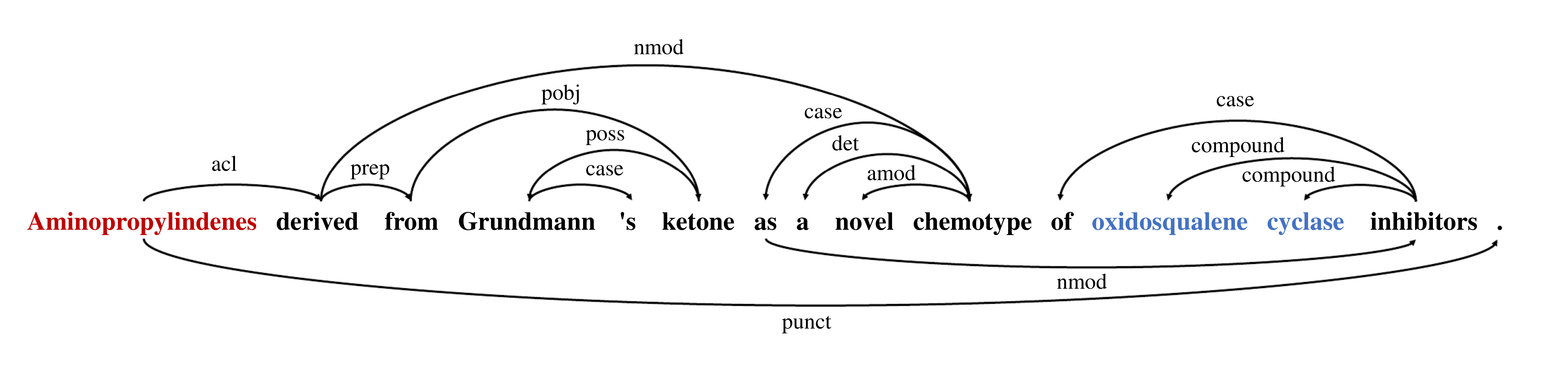}
\vskip -0.25in
\caption{1-best dependency tree for a biological sentence generated by the parser. \textbf{\textit{Aminopropylindenes}} and \textbf{\textit{oxidosqualene cyclase}} are the entities in the sentence.}
\label{dp-tree}
\vskip -0.15in
\end{figure*}

Medical relation extraction (MRE) refers to identifying relations among entities from medical literature and reports. It plays a very important role in downstream tasks such as medical knowledge graph construction~\cite{li2020real,rotmensch2017learning} and biomedical knowledge discovery~\cite{quirk2016distant}. On the other hand, as the number of medical literature increases, it becomes increasingly important to automatically discover the relation among entities in the literature~\cite{peng2017cross}. 

The addition of syntactic structure has been demonstrated to be beneficial for various natural language processing tasks~\cite{zaremoodi2017incorporating,zhou2005exploring,le2015forest}. As a type of syntactic structure, the dependency tree capturing long-distance connections between words can indeed improve benchmark relation extraction methods~\cite{tian2021dependency,chen2021relation,zhang2018graph,sun2020relation}. We demonstrate an example in Figure \ref{dp-tree}. Specifically, the 1-best dependency tree of the sentence ``\textbf{\textit{Aminopropylindenes}} derived from Grundmann's ketone as a novel chemotype of \textbf{\textit{oxidosqualene cyclase}} inhibitors'' in the CPR dataset. \textbf{\textit{Aminopropylindenes}} and \textbf{\textit{oxidosqualene cyclase}} are the entities, and the relation between them is “down regulator”, denoted as “CPR:4”.

However, in the medical field, the quality of the 1-best dependency tree generated by the out-of-domain parsers, e.g., parsers for the news domain, is relatively deficient. Generally, the main verb in a sentence is treated as the root node in the dependency tree, while, as the example shown in Figure \ref{dp-tree}, the entity \textbf{\textit{Aminopropylindenes}}, apparently a \textit{noun}, is treated as the root node. To solve this problem, multiple methods with dependency forests have been proposed~\cite{song2019leveraging, jin2020relation, guo2021learning}. Such approaches focus on redesigning the parser or substituting the parser with a semantic encoder, but the semantic and dependency tree syntactic information is used in a biased manner. Furthermore, the causality between the edges in the dependency forests and the performance of the model is not explored by benchmark methods.

To this end, we propose a novel approach, namely \textbf{C}ausality-\textbf{P}runed semantic dependency forest \textbf{G}raph \textbf{C}onvolutional \textbf{N}etwork (CP-GCN). To acquire the dependency forests enriched with semantic and syntactic information in an unbiased manner, we first obtain the 1-best dependency tree, as the sentence syntactic information, which is generated by the out-of-domain parser, and then fuse the syntactic information with the semantic information by using a switch gate network. The semantic information is captured in different representation subspaces using multi-head attention~\cite{vaswani2017attention}. To extract dependency forests' edges that are causally related to the MRE performance, we construct a causal explanation dataset based on Granger causality~\cite{granger1969investigating,granger1980testing} and train a task-specific causal explainer. We then obtain task-specific explanations of the dependency forests generated by the trained explainer and prune the dependency forests by following the corresponding explanations, which aim to eliminate the task-irrelevant information from the dependency forests. The pruned dependency forests are encoded by DCGCNs 
~\cite{guo2019densely} for MRE task. Empirically, the comparisons demonstrate that CP-GCN achieves state-of-the-art on benchmark relation extraction tasks, e.g., for the sentence-level relation extraction task, our model obtains 67.3 and 92.9 scores on CPR and PGR, respectively. The \textbf{contributions} are summarized as follows:
\begin{itemize}
    \item We propose an approach to generate dependency forests enriched with semantic and syntactic information in an unbiased manner.
    \item We propose a causal pruning approach to remove task-irrelevant information from the dependency forests, which is achieved by using a task-specific explainer trained on a causal explanation dataset for the target MRE task.
    \item CP-GCN achieves state-of-the-art on benchmark MRE datasets, and the ablation comparisons further support the effectiveness of each part of our model.
\end{itemize}

\section{Related Work}

\subsection{Medical Relation Extraction}
Previous work performs the MRE task by constructing the 1-best dependency tree of sentences~\cite{peng2017cross,song2018n}. However, the accuracy of the 1-best dependency tree generated by the out-of-domain parser is relatively low, resulting in a fall in MRE performance. Therefore,~\cite{song2019leveraging} proposes to use dependency forests to solve this problem, which uses EDGEWISE and KBESTEISNER algorithm to pick edges to construct dependency forests.~\cite{jin2020relation} encodes all effective dependency trees generated by a parser into dependency forests.~\cite{guo2021learning} utilizes multi-head attention and Kirchhoff’s Matrix-Tree Theorem (MMT)~\cite{koo2007structured} to automatically generate latent dependency forests without the usage of any parser. In general, ~\cite{song2019leveraging} and~\cite{jin2020relation} focus more on the syntactic information in the 1-best dependency tree generated by the out-of-domain parser, while~\cite{guo2021learning} directly discards the syntactic information and focuses only on the semantic information.

\subsection{Causal Explanation}
Causal explanation is designed to explain the importance of each module in a machine learning model on the prediction, which receives increasing attention recently~\cite{datta2016algorithmic,schwab2019cxplain,lin2021generative}. There are several viable forms of causality, including Granger causality~\cite{granger1969investigating}, causal Bayesian networks~\cite{pearl1985bayesian}, and structural causal models~\cite{pearl2009causality}.~\cite{chattopadhyay2019neural} proposes an attribution method based on the first principles of causality.~\cite{schwab2019cxplain} models the explanation task of image deep learning models as a causal learning task and proposes a causal explanation model based on Granger causality.~\cite{lin2021generative} proposes a framework for explaining graph neural networks using the first principles of Granger causality.

\section{Preliminaries}
\subsection{Task Definition}
Our task is to extract relation between entities in a sentence, focusing on both binary relation extraction and ternary relation extraction. Formally, the input to our task is a sentence $\boldsymbol{S} = \{w_1,w_2,\ldots,w_n\}$  with $n$ words and $w_i$ denotes the $i$-th word in the sentence. $\boldsymbol{S}$ is annotated with entity mentions \bm{$E_1$} and \bm{$E_2$}\footnote{\bm{$E_1$}, \bm{$E_2$} and \bm{$E_3$} for ternary relation extraction.}. The output is the relation between entities from a predefined relation set $\bm{R} =\{r_1,r_2,\ldots,r_m\}$, where 
$m$ denotes the number of relations.

\subsection{Densely-Connected Graph Convolutional Networks}
Graph Neural Network is a set of models that can effectively encode the information of graph structure, the classical models including Graph Attention Networks (GATs)~\cite{velickovic2017graph}, Graph Convolutional Networks (GCNs)~\cite{kipf2016semi}, etc. Densely-Connected Graph Convolutional Networks (DCGCNs)~\cite{guo2019densely} is a variant of GCNs, which introduces dense connections to GCNs. Thus being able to build multi-layer GCNs models with a large depth and learn richer information than the shallower GCNs models. More specifically, DCGCNs differs from GCNs in that the embedding of node $v$ in the $l$-th layer receives information from all the preceding layers, which can be formulated as follows:
\begin{equation}
    \bm{h}^{(l)}_v=\rho\left(\sum_{u \in \mathcal{N}(v)}\bm{W}^{(l)} \times \bm{g}^{(l)}_u+\bm{b}^{(l)}\right)
    \label{d1}
\end{equation}
where $\times$ denotes matrix multiplication, $\bm{h}^{(l)}_v$ is the embedding of node $v$ in the $l$-th layer, $\rho$ is an activation function, $\mathcal{N}(v)$ denotes the neighbours of node $v$, $\bm{W}^{(l)}$ and $\bm{b}^{(l)}$ are the weight matrix and bias vector of the $l$-th layer respectively, and $\bm{g}^{(l)}_u$ indicates the information about node $u$ from all the preceding layers. Mathematically, $\bm{g}^{(l)}_u$ can be calculated by concatenating the initial embedding $\bm{x}_u$ and the node embedding $\bm{h}^{(1)}_u;\ldots;\bm{h}^{(l-1)}_u$ produced in layer $1,\ldots,l-1$, respectively. 
\begin{equation}
    \bm{g}^{(l)}_u=[\bm{x}_u;\bm{h}^{(1)}_u;\ldots;\bm{h}^{(l-1)}_u] 
    \label{d2}
\end{equation}

\subsection{Dependency Tree Generation}
To construct the 1-best dependency tree, we use Standard CoreNLP Toolkits (SCT)~\cite{DBLP:conf/acl/ManningSBFBM14} to obtain the dependency tree $\mathcal{T}$ for each input sentence $\boldsymbol{S}$ and represent $\mathcal{T}$ by a adjacency matrix $\mathbf{T} = (t_{i,j})_{n \times n}$\footnote{The adjacency matrix $\mathbf{T}$ adds the self-loop of each word to the dependency tree $\mathcal{T}$ with the ``self'' dependency type and regards the dependencies between words as unoriented.}, where $t_{i,j}$ is the dependency type (e.g., \emph{dobj}) between $w_i$ and $w_j$, e.g., $t_{i,j} = 0$ if the connection between $w_i$ and $w_j$ do \textit{not} exist. Then, we encode $t_{i,j}$ to the corresponding embedding $c^t_{i,j}$ with a learnable matrix, and use $\mathbf{C} = (c^t_{i,j})_{n \times n}$ to denote the syntactic matrix.

\section{Methodology}
\begin{figure*}[t]
\centering
\includegraphics[scale=0.25]{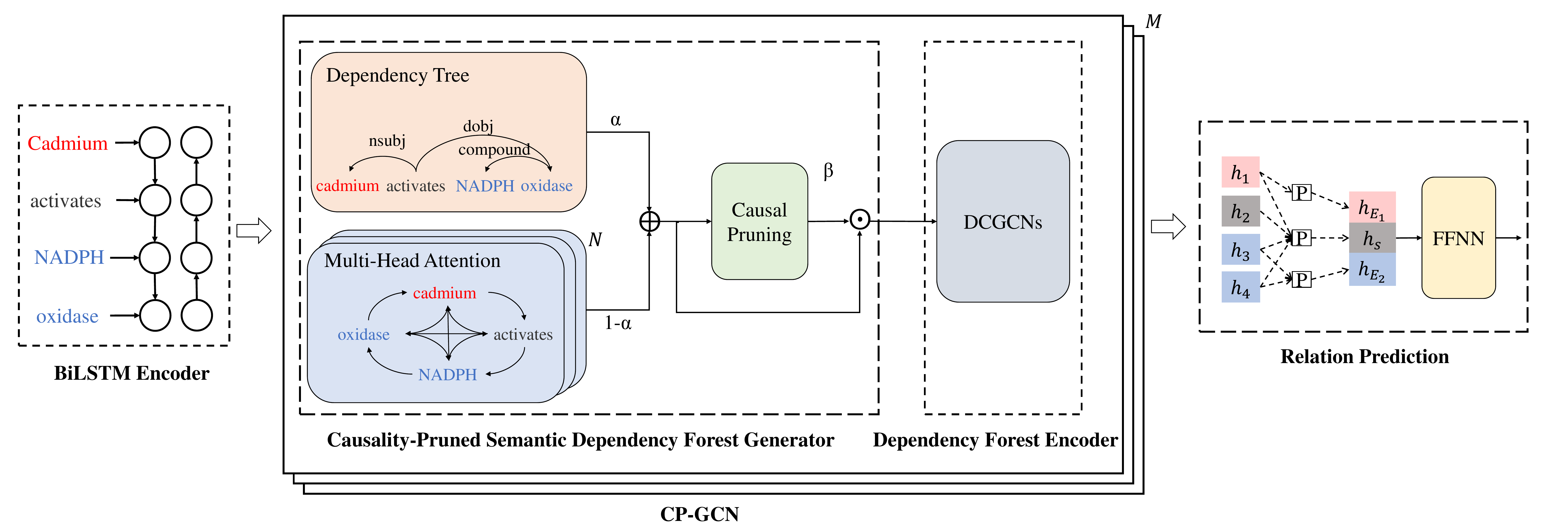}
\vskip -0.15in
\caption{The overall architecture of CP-GCN with an example input sentence (``\textbf{\textit{Cadmium}}'' in red and ``\textbf{\textit{NADPH oxidase}}'' in blue are two entities of the sentence). The model consists of three components: 1) BiLSTM Encoder obtains the sentence representation with a BiLSTM model. 2) CP-GCN is the main component of the model which contains $M$ identical blocks, and each block contains two modules. Causality-Pruned Semantic Dependency Forest Generator combines the dependency tree and the multi-head attention with $N$ heads to generate dependency forests and then prunes them using a task-specific causal explainer. Dependency Forest Encoder uses DCGCNs to encode the pruned dependency forests. 3) Relation Prediction module predicts relations using global and local max pooling and feedforward neural networks (FFNN).}
\label{overall}
\vskip -0.1in
\end{figure*}

In this section, we introduce our proposed CP-GCN model shown in Figure \ref{overall}.
\subsection{Causality-Pruned Semantic Dependency Forest Generator}
In the medical domain, the quality of the 1-best dependency tree generated by the out-of-domain parsers is relatively deficient. Thus, we propose the Causality-pruned Semantic dependency Forest Generator (CSFG) to generate dependency forests enriched with syntactic and semantic information and derive task-relevant information from them.
\subsubsection{Semantic Embedding Module}
In order to construct dependency forests that combine both semantic and syntactic information in an unbiased manner, we propose a semantic embedding module to incorporate the semantic information of the sentence into the 1-best dependency tree.  

Specifically, we model semantic information using the multi-head attention mechanism~\cite{vaswani2017attention} with $N$ heads, which captures the semantic relevance between words in a sentence.  For the $p$-th head, we compute the semantic matrix $\mathbf{A}^p$ by using the query vector $\bm{Q}$ and the key vector $\bm{K}$:
\begin{equation}
    \mathbf{A}^p = \frac{\left(\bm{Q} \times \bm{W}^{\bm{Q}}\right) \times \left(\bm{K} \times \bm{W}^{\bm{K}}\right)^\top}{\sqrt{d}}
    \label{a1}
\end{equation}
where $\bm{W}^{\bm{Q}}$ and $\bm{W}^{\bm{K}}$ are learnable transformation matrices for $\bm{Q}$ and $\bm{K}$, respectively, and $d$ is the dimension of $\bm{K}$.

Finally, the $p$-th dependency forest can be obtained by summing the syntactic matrix $\mathbf{C}$ and the semantic matrix $\mathbf{A}^p$ with a switch gate network and a softmax function:
\begin{equation}
    \mathbf{F}^p = \textrm{softmax}\left(\left(1-\alpha\right)\,\mathbf{A}^p + \alpha\, \mathbf{C}\right)
    \label{a2}
\end{equation}
where $\alpha \in [0,1]$ is a hyper-parameter to balance the syntactic matrix $\mathbf{C}$ and the semantic matrix $\mathbf{A}^p$, and  $\mathbf{F}^p$ is the adjacency matrix of the $p$-th dependency forest.
\subsubsection{Task-Specific Causal Pruning Module}

\begin{figure*}[t]
\centering
\includegraphics[scale=0.2]{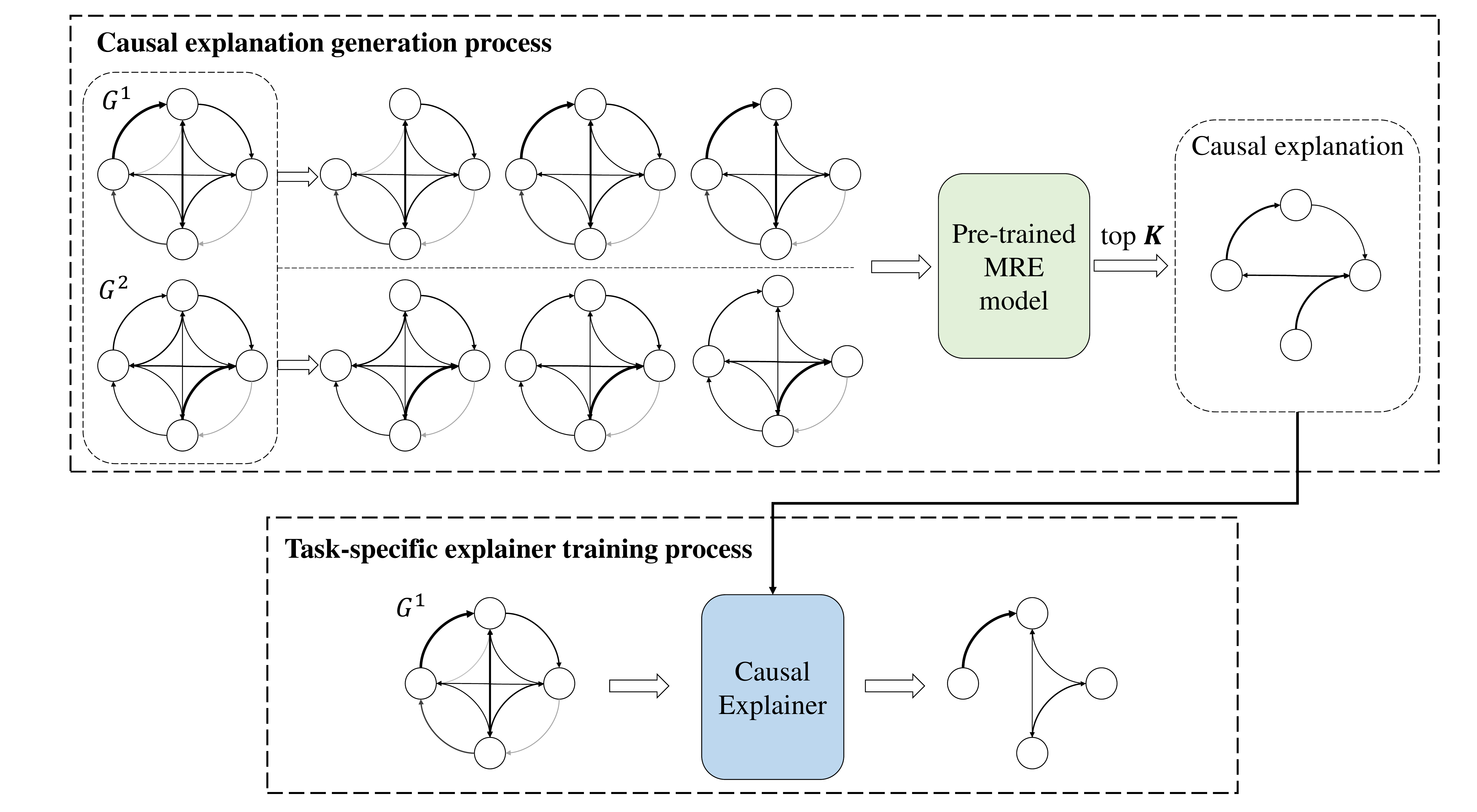}
\vskip -0.1in
\caption{Illustration of training a task-specific causal explainer. Causal explanation generation process generates the causal explanations for the dependency forests using a pre-trained MRE model and the designed rules. Task-specific explainer training process trains a task-specific causal explainer with the \textit{generated} causal explanation dataset.}
\label{causal}
\vskip -0.1in
\end{figure*}

In this part, our major objective is to extract dependency forests' edges that are causally related to the MRE performance. Inspired by \cite{lin2021generative}, we propose a method consisting of three processes for pruning dependency forests based on Granger causality. The first two processes aim to train a task-specific causal explainer, which are illustrated in Figure \ref{causal}. The causal pruning process prunes the dependency forests with the trained causal explainer.

\textbf{Causal explanation generation process.} This process is designed to construct a causal explanation dataset for a specific MRE task. Given a pre-trained MRE model denoted by $f_{\textrm{MRE}}(\cdot)$ and the gold-standard relation $r$ of the sentence $\boldsymbol{S}$. We start by using the semantic embedding module of the pre-trained MRE model to generate $N \ast M$ dependency forests of the sentence $\boldsymbol{S}$, denoted by $\mathcal{G}=\{\bm{G}^1,\bm{G}^2,\ldots,\bm{G}^{N \ast M}\}$. For any dependency forest $\bm{G}^i$, it can be represented as $\bm{G}^i=(\mathbf{F}^i,\mathbf{H}_0)$, where $\mathbf{F}^i$ is the fully-connected adjacency matrix indicating the weights of the edges, and $\mathbf{H}_0$ is the matrix of node features, which is the same for each dependency forest. Then, we need to extract the top $\mathcal{K}$ edges from the dependency forests that are most relevant for predicting relation $r$. We implement this based on Granger causality.\footnote{Granger causality describes the causal relationships between two (or more) variables. Specifically, if we are better able to predict variable $\widetilde{y}$ using all information U than excluding information about variable $x$, which means that the variable $x$ helps predict variable $\widetilde{y}$. Then we say that $x$ Granger-causes $\widetilde{y}$~\cite{granger1980testing}, denoted by $x \rightarrow \widetilde{y}$.}

Specifically, we use $\mathcal{L}_{\mathcal{G}}$ to denote the model error of $f_{\textrm{MRE}}(\cdot)$ when taking the $N \ast M$ dependency forests into account, and $\mathcal{L}_{\mathcal{G} \setminus \{e_k\}}$ represents the model error excluding the edge $e_k$ from each dependency forest. According to Granger causality, we can quantify the causal contribution of edge $e_k$ to our MRE task by the change in model error after removing edge $e_k$: 
\begin{equation}
    \Delta_{e_k} = \mathcal{L}_{\mathcal{G} \setminus \{e_k\}}-\mathcal{L}_{\mathcal{G}}
    \label{c1}
\end{equation}
where $\Delta_{e_k}$ represent the causal contribution of edge $e_k$.

To calculate  $\mathcal{L}_{\mathcal{G}}$ and $\mathcal{L}_{\mathcal{G} \setminus \{e_k\}}$, we first take the $N \ast M$ dependency forests $\mathcal{G}$ and $\mathcal{G} \setminus \{e_k\}$ as the input to the pre-trained model, respectively, and obtain their corresponding outputs $r_{\mathcal{G}}$ and $r_{\mathcal{G} \setminus \{e_k\}}$:
\begin{align}
    r_{\mathcal{G}}&=f_{\textrm{MRE}}\left(\mathbf{F}^1, \ldots, \mathbf{F}^{N \ast M}\right)\\
    r_{\mathcal{G} \setminus \{e_k\}}&=f_{\textrm{MRE}}\left(\mathbf{F}^1 \setminus \{e_k\}, \ldots, \mathbf{F}^{N \ast M} \setminus \{e_k\}\right)
\end{align}

We then use the cross-entropy loss function to measure the model error, denoted as \textbf{CE}.
\begin{align}
    \mathcal{L}_{\mathcal{G}}&=\textbf{CE}\left(r,r_{\mathcal{G}}\right)\\
    \mathcal{L}_{\mathcal{G} \setminus \{e_k\}}&=\textbf{CE}\left(r,r_{\mathcal{G} \setminus \{e_k\}}\right)
\end{align}

Finally, we filter out the edges with the top $\mathcal{K}$ causal contributions to form the causal explanation. In summary, our causal explanation dataset is constructed with dependency forests and the corresponding causal explanations. Therefore, such a dataset is relevant to the specific MRE task.

\textbf{Task-specific explainer training process.} This process generates a task-specific explainer based on the causal explanation dataset. Following~\cite{lin2021generative}, we use an encoder-decoder architecture as the explainer. The encoder consists of several graph convolutional layers to aggregate information between neighbors in the dependency forest and learn node features. The decoder uses the inner product operation to obtain the explanation matrix. Specifically, the explanation matrix for $\bm{G}^i$ can be obtained by the explainer as:
\begin{align}
    \label{c6} \mathbf{X}^i&=\sigma\left(f_{\textrm{GCN}}\left(\mathbf{F}^i,\mathbf{H}_0\right) \times f_{\textrm{GCN}}\left(\mathbf{F}^i,\mathbf{H}_0\right)^\top\right)
\end{align}
where $f_{\textrm{GCN}}(\cdot)$ denotes graph convolutional layers, $\mathbf{X}^i$ is the explanation matrix and each value in $\mathbf{X}^i$ represents the contribution of its corresponding edge to the prediction relation $r$, and $\sigma$ is the activation function.
\textbf{Causal pruning process.} Based on the pre-trained explainer, task-relevant explanation of the dependency forest can be obtained. Given the $\mathbf{F}^p$ calculated by Eq. \ref{a2} and the pre-trained explainer, the explanation matrix $\mathbf{X}^p$ corresponding to $\mathbf{F}^p$ can be calculated via Eq. \ref{c6}. Causal pruning for $\mathbf{F}^p$ can be formulated as:
\begin{equation}
    \mathbf{\hat{F}}^p = \textrm{softmax}\left(\mathbf{F}^p \odot \left(1 + \beta\, \mathbf{X}^p\right)\right) 
\end{equation}
where $\odot$ is the element-wise multiplication, and $\beta \in [0,1]$ is a hyper-parameter to control the coefficient of the explanation matrix $\mathbf{X}^p$.

\subsection{Dependency Forest Encoder}
Given $N$ pruned dependency forests, DCGCNs are used to encode information from the forest structure. For the $p$-th pruned dependency forest, which is represented by the adjacency matrix $\mathbf{\hat{F}}^p$. We use DCGCNs with $L$ layers to aggregate information about neighbors in $\mathbf{\hat{F}}^p$, and the representation of node $i$ at the $l$-th layer can be calculated as:
\begin{equation}
    \bm{h}^{(l)}_{p_i}=\rho\left(\sum_j^n\mathbf{\hat{F}}^p_{ij}\left(
    \bm{W}^{(l)}_p \times \bm{g}^{(l)}_{p_j}+\bm{b}^{(l)}_p\right)\right)
\end{equation}
where $\mathbf{\hat{F}}^p_{ij}$ denotes the weight between node $i$ and node $j$ in $\mathbf{\hat{F}}^p$. $\bm{g}^{(l)}_{p_j}$ denotes the information about node $j$ in the $p$-th pruned dependency forest from all the preceding layers and can be obtained by the same way as Eq. \ref{d2}.

Then, we concatenate the representations obtained from the $N$ dependency forests and fuse them using a linear layer. This process can be formulated as follows:
\begin{align}
    \mathbf{H}_{\textrm{b}}&= \textrm{Linear}\left([\mathbf{\bm{H}}^1;\mathbf{\bm{H}}^2;\ldots;\mathbf{\bm{H}}^N]\right)
\end{align}
where $\mathbf{\bm{H}}^i$ is the node representations obtained by DCGCNs for the $i$-th dependency forest, and $\mathbf{H}_{\textrm{b}}$ denotes the node representations of each block. $M$ identical blocks are combined in the same way as above to obtain the final node representations for sentence $\boldsymbol{S}$, denoted as $\mathbf{H}$.
\subsection{Relation Prediction}
To predict the relations among entities, the max pooling mechanism is used. We obtain the global sentence representation $\bm{h}_{\boldsymbol{S}}$ by applying the max pooling function to all the words in sentence $\boldsymbol{S}$:
\begin{align}
    \bm{h}_{\boldsymbol{S}} = \textrm{MaxPooling}(\{\bm{h}_1,\ldots,\bm{h}_n\})
\end{align}
where $\bm{h}_i$ is the feature vector of word $w_i$, and then obtain the representation of each entity by applying the max pooling function to the words that belongs to an entity mention (i.e., $\bm{E}_q$). Therefore, the entity representation of $\bm{E}_q$ can be obatined by:
\begin{align}
    \bm{h}_{\bm{E}_q} = \textrm{MaxPooling}(\{\bm{h}_i|w_i \in \bm{E}_q\})
\end{align}

The sentence representation and entity representations are concatenated and fed into a feed-forward neural network (FFNN), and then we transform it into an $m$-dimensional vector $\bm{h}_{\bm{R}}$ using a linear layer to make a prediction:
\begin{align}
    \bm{h}_{\bm{R}} = \textrm{Linear}\left(\textrm{FFNN}\left([\bm{h}_{\boldsymbol{S}};\bm{h}_{\bm{E}_1};\ldots;\bm{h}_{\bm{E}_{\mathcal{Q}}}]\right)\right)
\end{align}
where $\mathcal{Q}$ is 2 in the binary relational extraction task and is 3 in the ternary relational extraction task, $m$ denotes the number of relations $\bm{R}$.

\section{Experiment}
\subsection{Datasets}
\begin{table}
\small
\centering
\setlength{\tabcolsep}{15pt}
\renewcommand{\arraystretch}{1.2}
\begin{tabular}{lrrr}
\hline
&CPR&PGR\\
\hline
TRAIN&16107&11780\\
DEV&10030&-\\
TEST&14269&219\\
\hline
\end{tabular}
\vskip -0.1in
\caption{\label{number}
The number of instances of CPR and PGR.
}
\vskip -0.15in
\end{table}
We evaluate our model on three datasets with two types of tasks: cross-sentence n-ary relation extraction and sentence-level relation extraction following~\cite{guo2021learning}. 

For the cross-sentence n-ary relation extraction task, we use the dataset extracted by~\cite{peng2017cross} based on PubMed. Most of the instances in this dataset contain multiple sentences, and the entities in the instances are cross-sentence. In detail, this dataset contains 6987 instances of ternary relations and 6087 instances of binary relations, each of them is divided into five folders according to \cite{song2018n}. The relation between entities in each instance belongs to one of the relation sets, including ``resistance or non-response'', ``sensitivity'', ``response'', ``resistance'', and ``None''. Following \cite{guo2021learning}, we define two sub-tasks on this dataset: multi-class and binary-class relation extraction. For multi-class relation extraction, we keep the original dataset unchanged, and for binary-class relation extraction, we define the first four relations as ``Yes'' and the ``None'' as ``No''. 

For the sentence-level relation extraction task, we use two datasets for Medical Relation Extraction, namely, BioCreative Vi CPR (CPR)~\cite{krallinger2017overview} and Phenotype-Gene relation (PGR)~\cite{sousa2019silver}. CPR focuses on the relations between chemical components and human proteins, which contains six relation types (``CPR:3'', ``CPR:3'', ``CPR:4'', ``CPR:5'', ``CPR:6'', ``CPR:9'', ``None''). PGR focuses on whether human phenotypes are related to human genes, which contains two relation types (``TRUE'' for related and ``FALSE'' for unrelated). The number of instances for train/dev/test sets of CPR and PGR datasets is shown in Table \ref{number}.

\subsection{Implementation}
During the causal explanation generation process, we use a pre-trained CP-GCN model \textit{without} the task-specific causal pruning module as $f_{\textrm{MRE}}(\cdot)$ and choose 1/5 of the training set for the~\cite{peng2017cross} dataset while the full training set for other datasets to generate the full dependency forests. Then, we set $\mathcal{K}=20$ to construct causal explanation datasets.

For evaluation, we follow previous studies to use the test accuracy averaged over five cross validation folds for the cross-sentence n-ary task and F1 scores for the sentence-level task. Refer to the supplementary files for the details.

See Appendix \ref{he} for the hyper-parameter experiment on $N$, $\alpha$, and $\beta$.

\subsection{Results on Cross-Sentence N-Ary Relation Extraction Task}
\begin{table*}
\centering
\resizebox{\textwidth}{!}{
\begin{tabular}{clcccccccc}
\hline
\multirow{3}{*}{Syntax Type}& \multirow{3}{*}{Model}& \multicolumn{4}{c}{Binary-class} & \multicolumn{4}{c}{Multi-class}\\
\cline{4-5}
\cline{8-9}
& &\multicolumn{2}{c}{Ternary}&\multicolumn{2}{c}{Binary}&\multicolumn{2}{c}{Ternary}&\multicolumn{2}{c}{Binary} \\
\cline{3-6}
\cline{7-10}
& &\multicolumn{1}{c}{Single}&\multicolumn{1}{c}{Cross}&\multicolumn{1}{c}{Single}&\multicolumn{1}{c}{Cross}&\multicolumn{1}{c}{Single}&\multicolumn{1}{c}{Cross}&\multicolumn{1}{c}{Single}&\multicolumn{1}{c}{Cross} \\
\hline
\multirow{4}{*}{Tree}&DAG LSTM~\cite{peng2017cross}&77.9&80.7&74.3&76.5&-&-&-&-\\
&GRN~\cite{song2018n}&80.3&83.2&83.5&83.6&-&71.7&-&71.7\\
&GCN(Full)~\cite{zhang2018graph}&84.3&84.8&84.2&83.6&-&77.5&-&74.3\\
&GCN(Pruned)~\cite{zhang2018graph}&85.8&85.8&83.8&83.7&-&78.1&-&73.6\\
\hline
\multirow{5}{*}{Forest}&AGGCN~\cite{guo2019attention}&87.1&87&85.2&85.6&-&79.7&-&77.4\\
&AGGCN*~\cite{guo2019attention}&86.3&87.2&86.3&85.8&77.7&78.7&77.7&77.3\\
&LF-GCN~\cite{guo2021learning}&88&88.4&86.7&\textbf{87.1}&-&81.5&-&79.3\\
&LF-GCN*~\cite{guo2021learning}&88.2&88.3&87&86.3&82.9&83.9&80&79.6\\
&AC-GCN~\cite{qian2021auto}&88.8&88.8&86.8&86.5&-&84.6&-&\textbf{81}\\
\hline
&CP-GCN(ours)&\textbf{89.5}&\textbf{89.1}&\textbf{87.3}&86.5&\textbf{84.3}&\textbf{84.9}&\textbf{81}&80.1\\
\hline
\end{tabular}
}
\vskip -0.1in
\caption{\label{cross-sentence}
Average test accuracies on the~\cite{peng2017cross} dataset for binary-class n-ary relation extraction and multi-class n-ary relation extraction. ``Ternary'' denotes drug-gene-mutation tuple and ``Binary'' denotes drug-mutation pair. ``Single'' means considering the instances within a single sentence, while ``Cross'' means considering all instances. Models with * indicate the accuracy of our reimplementation on their released implementation.
}
\vskip -0.1in
\end{table*}
For the cross-sentence n-ary relation extraction task, We compare CP-GCN against two kinds of models and report the average test accuracies on the~\cite{peng2017cross} dataset in table \ref{cross-sentence}.

\textbf{Tree}: models use the 1-best dependency tree. DAG LSTM, GRN, and GCN(Full) use the full dependency tree directly, while GCN(Pruned) generates a pruned dependency tree with some rules~\cite{zhang2018graph}. Besides, DAG LSTM uses graph-structure LSTM to encode the dependency tree, while GRN and GCN use graph recurrent networks and graph convolutional networks, respectively. 

\textbf{Forest}: models construct dependency forests. ACGCN treats a fully connected graph obtained by multi-head attention as a forest. LF-GCN automatically generates latent forests using multi-head attention and MMT. AC-GCN generates dependency forests with multi-head attention and encodes them with a 2D convolutional network.

As shown in Table \ref{cross-sentence}, our proposed CP-GCN model achieves state-of-the-art performance in most settings. Specifically, the model using the pruned dependency tree performs better than those using the full dependency tree, suggesting that noisy information does exist in the 1-best dependency tree. In addition, the forest structure shows an advantage on this task, while CP-GCN surpasses the current state-of-the-art forest-structured model (AC-GCN) by  0.7 and 0.3 points on the binary-class ternary relation extraction task. The multi-class n-ary relation extraction task in~\cite{peng2017cross} dataset is more challenging due to the unbalanced distribution of each relation, and CP-GCN can consistently achieve comparable performance.

\subsection{Results on Sentence-Level Relation Extraction Task}
For the sentence-level relation extraction task, we implement our approach on the CPR and PGR datasets and compare it against state-of-the-art models. We classify these models into three groups according to their syntax type.

\textbf{None}: models do not use tree or forest structures. Att-GRU adds a self-attention layer to GRU, and Bran uses a bi-affine self-attention model to capture interactions in sentences. BioBERT is a biomedical pre-trained language representation model.

\textbf{Tree}: models use the 1-best dependency tree. GCN, Tree-DDCNN, and Tree-GRN encode the full tree with GCN, DDCNN, and GRN, respectively. BO-LSTM prunes the tree, retaining only the shortest dependency path.

\textbf{Forest}: models construct dependency forests. Edgewise-GRN chooses edges with weights greater than the pre-defined threshold to form the dependency forest. KBest-GRN constructs the forest by aggregating K-best trees. ForestFT-DDCNN generates forests with a learnable dependency parser.

\begin{table}
\centering
\resizebox{\columnwidth}{!}{
\begin{tabular}{clc}
\hline
Syntax Type&Model&F1\\
\hline
\multirow{2}{*}{None}&Att-GRU~\cite{liu2017attention}&49.5\\
&Bran~\cite{verga2018simultaneously}&50.8\\
\hline
\multirow{3}{*}{Tree}&GCN~\cite{zhang2018graph}&52.2\\
&Tree-DDCNN~\cite{jin2020relation}&50.3\\
&Tree-GRN~\cite{jin2020relation}&51.4\\
\hline
\multirow{5}{*}{Forest}&Edgewise-GRN~\cite{song2019leveraging}&53.4\\
&KBest-GRN~\cite{song2019leveraging}&52.4\\
&AGGCN~\cite{guo2019attention}&56.7\\
&ForestFT-DDCNN~\cite{jin2020relation}&55.7\\
&LF-GCN~\cite{guo2021learning}&58.9\\
&AC-GCN~\cite{qian2021auto}&65.8\\
\hline
&CP-GCN(ours)&\textbf{67.3}\\
\hline
\end{tabular}
}
\vskip -0.1in
\caption{\label{cpr}
Main results on CPR.
}
\vskip -0.2in
\end{table}

The results of CPR and PGR datasets are shown in Table \ref{cpr} and Table \ref{pgr}. CP-GCN achieves state-of-the-art performance on both datasets. F1 score increases by 1.5 and 0.5 on the CPR and PGR datasets, respectively. Compared to models with forest structure, CP-GCN performs significantly better than both models with a bias towards syntactic information (Edgewise-GRN, KBest-GRN, and ForestFT-DDCNN) and models using almost only semantic information (AGGCN, LF-GAN, and AC-GCN), which demonstrates the effectiveness of our proposed CSFG method. 
\begin{table}
\centering
\resizebox{\columnwidth}{!}{
\begin{tabular}{clc}
\hline
Syntax Type& Model& F1\\
\hline
\multirow{1}{*}{None}&BioBERT~\cite{lee2020biobert}&67.2\\
\hline
\multirow{3}{*}{Tree}&BO-LSTM~\cite{lamurias2019bo}&52.3\\
&GCN~\cite{zhang2018graph}&81.3\\
&Tree-GRN~\cite{jin2020relation}&78.9\\
\hline
\multirow{5}{*}{Forest}&Edgewise-GRN~\cite{song2019leveraging}&83.6\\
&KBest-GRN~\cite{song2019leveraging}&85.7\\
&AGGCN~\cite{guo2019attention}&89.3\\
&ForestFT-DDCNN~\cite{jin2020relation}&89.3\\
&LF-GCN~\cite{guo2021learning}&91.9\\
&AC-GCN~\cite{qian2021auto}&92.4\\
\hline
&CP-GCN(ours)&\textbf{92.9}\\
\hline
\end{tabular}
}
\vskip -0.1in
\caption{\label{pgr}
Main results on PGR.
}
\vskip -0.1in
\end{table}

\subsection{Analysis and Discussion}
\begin{figure}[t]
\centering
\includegraphics[scale=0.6]{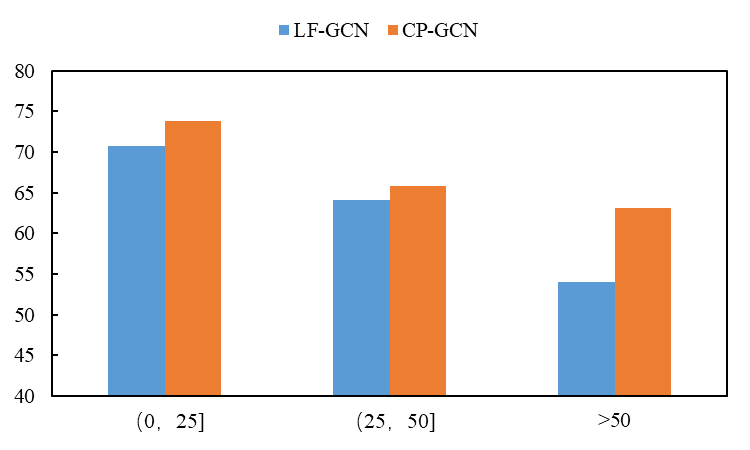}
\vskip -0.15in
\caption{F1 scores against sentence length. The results on LF-GCN are reproduced based
on its released implementation.}
\label{sl}
\vskip -0.1in
\end{figure}

\begin{figure}[t]
\centering
\includegraphics[scale=0.23]{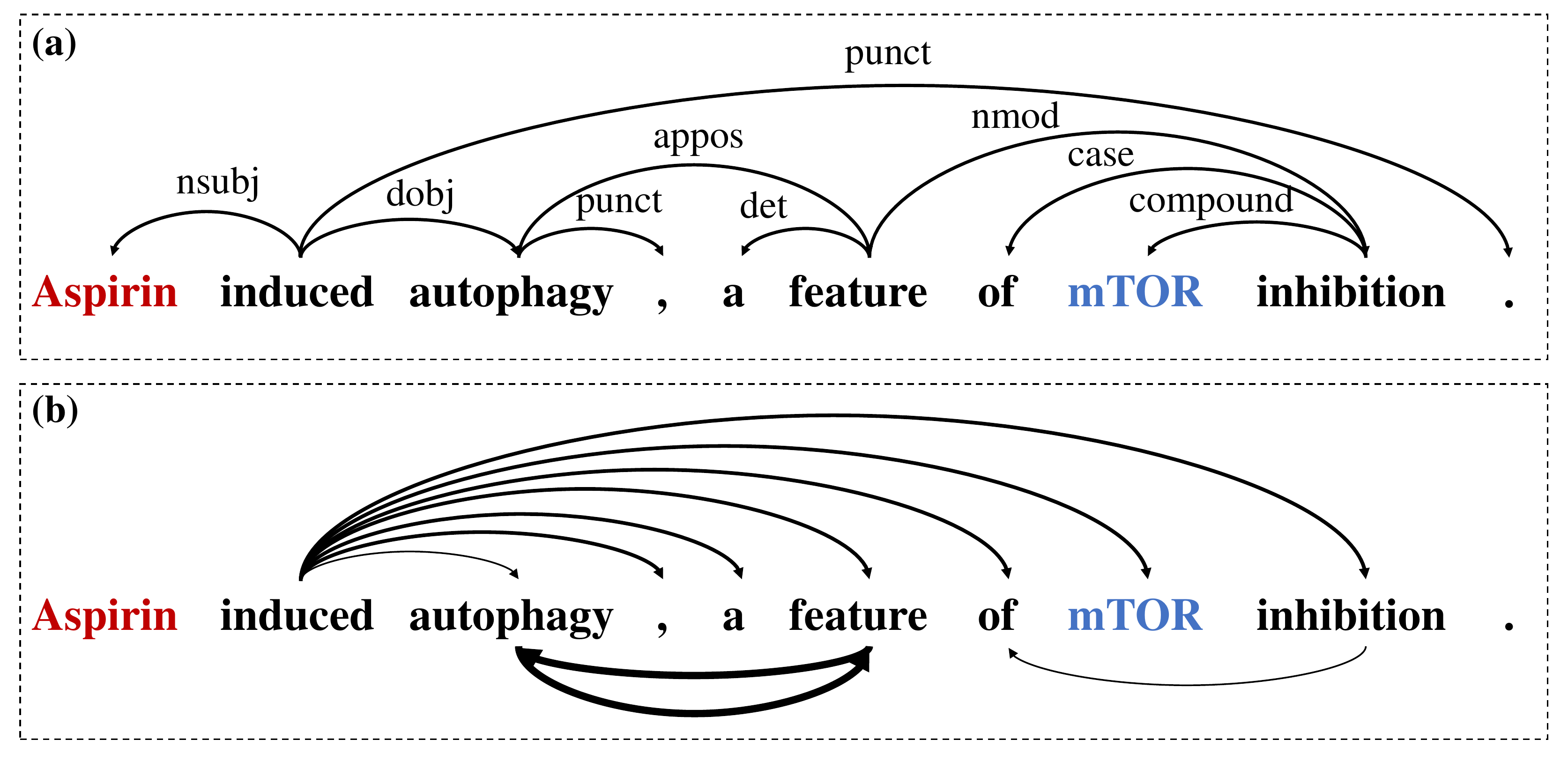}
\vskip -0.1in
\caption{Visualizations of (a) 1-best dependency tree and (b) top 10 highest causal weight edges of the pruned dependency forest for the example input, where thicker lines denote the connections with higher causal weights.}
\label{case}
\vskip -0.2in
\end{figure}

\textbf{Ablation study.}
To validate the effectiveness of the ingredients of CP-GCN, i.e., the semantic embedding module and the task-specific causal pruning module, we conduct the ablation study on CPR. We train the complete CP-GCN, an ablation model without the semantic embedding module, and another ablation model without the task-specific causal pruning module, respectively. Our experimental results are reported in Table \ref{ablation}. We observe that the performance of the model dropped (compared with complete CP-GCN) regardless of which module is removed, suggesting that both modules can help construct dependency forests that are more conducive to predicting relation. Comparing these two modules, the removal of the task-specific causal pruning module has a greater impact on performance, which suggests that the proposed causal pruning method can effectively distinguish vital information from noise.
\begin{table}
\centering
\small
\setlength{\tabcolsep}{15pt}
\renewcommand{\arraystretch}{1.2}
\begin{tabular}{lc}
\hline
Model& F1\\
\hline
CP-GCN&67.3\\
  -semantic embedding module&66.7\\
  -task-specific causal pruning module&65.7\\
\hline
\end{tabular}
\vskip -0.1in
\caption{\label{ablation}
An ablation study for CP-GCN on CPR dataset.
}
\vskip -0.25in
\end{table}
\\
\textbf{Performance against sentence length.}
Figure \ref{sl} compares the F1 scores of our CP-GCN model and the LF-GCN model~\cite{guo2021learning} under different sentence lengths. Following~\cite{guo2021learning}, We divide the test set of CPR into three groups ((0,25], (25,50], >50) based on sentence length. In general, CP-GCN outperforms LF-GCN against various sentence lengths. Otherwise, our model achieves a significant improvement on the more challenging long sentences, which demonstrates the ability of our model to capture long-range dependencies. Moreover, the dependency forests of the long sentences are more sophisticated, thus indicating that the task-specific causal explainer is able to extract task-relevant information from the sophisticated graph structure.
\\
\textbf{Case study.}
To further validate the efficiency of CP-GCN, we conduct a case study on an example sentence ``\textbf{\textit{Aspirin}} induced autophagy, a feature of \textbf{\textit{mTOR}} inhibition'', which can be correctly predicted by our model to be a ``down regulator'' relation between \textbf{\textit{Aspirin}}($\bm{E_1}$) and \textbf{\textit{mTOR}}($\bm{E_2}$). Figure \ref{case}(a) shows its 1-best dependency tree, and Figure \ref{case}(b) visualizes the top 10 edges with the highest causal weights in the pruned dependency forest generated by the proposed CSFG and the thicker lines referring to higher causal weights. In this example, the connection between ``\textit{induced}'' and ``\textit{feature}'' enhances in the pruned dependency forest, and we reckon the latent reason is that CP-GCN can capture richer semantic information. We observe that there exists a strong connection between ``\textit{autophagy}'' and ``\textit{feature}'' in the pruned dependency forest, which improves the prediction of the relation between \textbf{\textit{Aspirin}} and \textbf{\textit{mTOR}}, supporting the effectiveness of the task-specific causal explainer.

\section{Conclusion}
In this paper, we introduce a novel approach for the medical relation extraction task, namely CP-GCN, which proposes a causality-pruned dependency forest enriched with semantic and syntactic information. We first construct dependency forests by incorporating semantic information into the dependency tree generated by the off-the-shelf parser. Then, a task-specific causal explainer is trained to prune the dependency forests. Experiments on the benchmark medical datasets demonstrate the superiority of CP-GCN over the state-of-the-art methods for the medical relation extraction task.

\section*{Acknowledgements}
This work is supported by the National Key Research and Development Program of China (No. 2019YFB1405100).

\bibliography{custom}
\bibliographystyle{acl_natbib}

\newpage

\appendix

\section{Appendix}
\subsection{Hyper-Parameter Experiment}
\label{he}
\begin{figure*}[!htb]  
\subfigure[]{
 \label{fig:a}     
\includegraphics[width=0.3\textwidth]{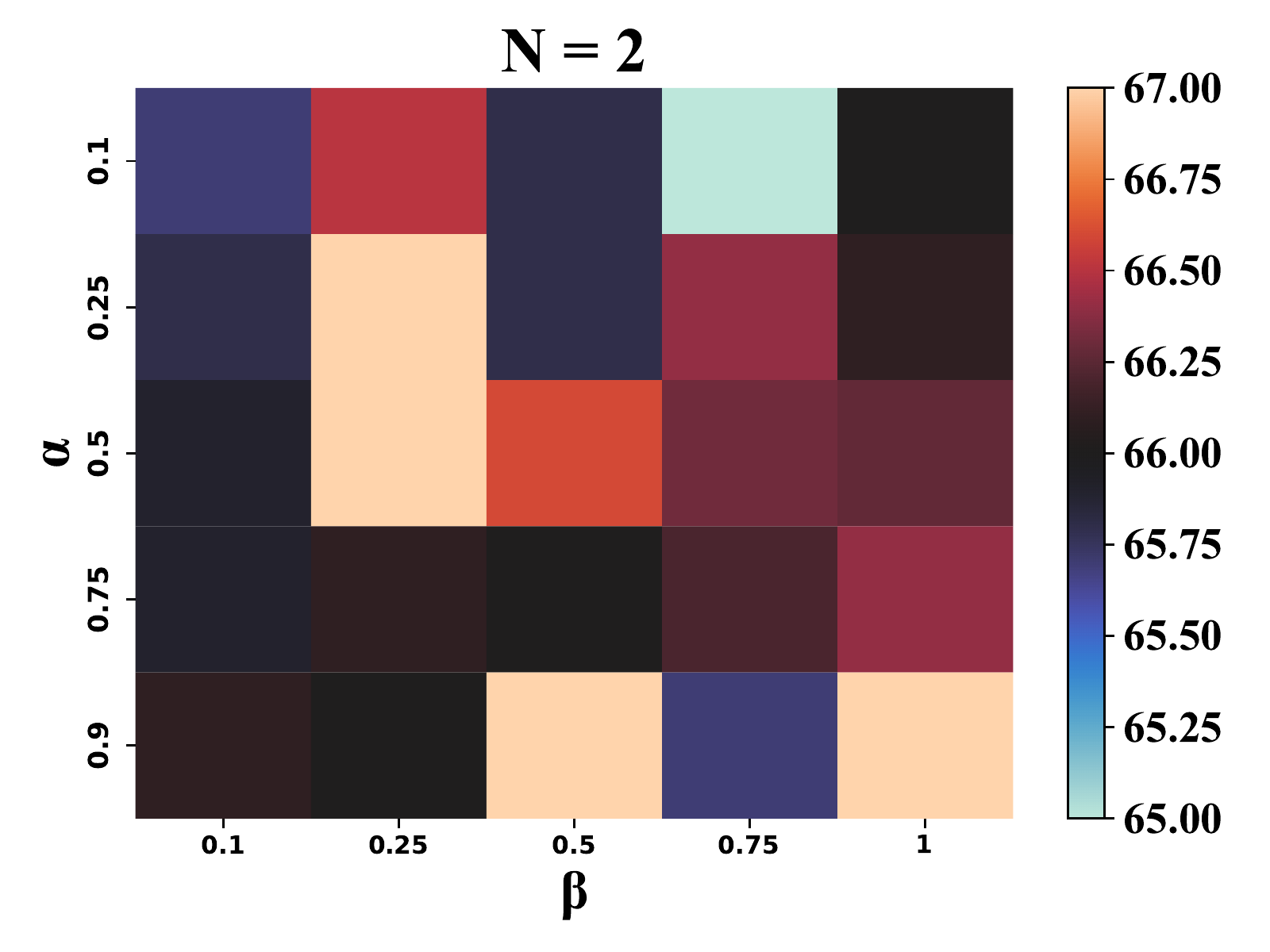}  
}     
\subfigure[]{ 
\label{fig:b}     
\includegraphics[width=0.3\textwidth]{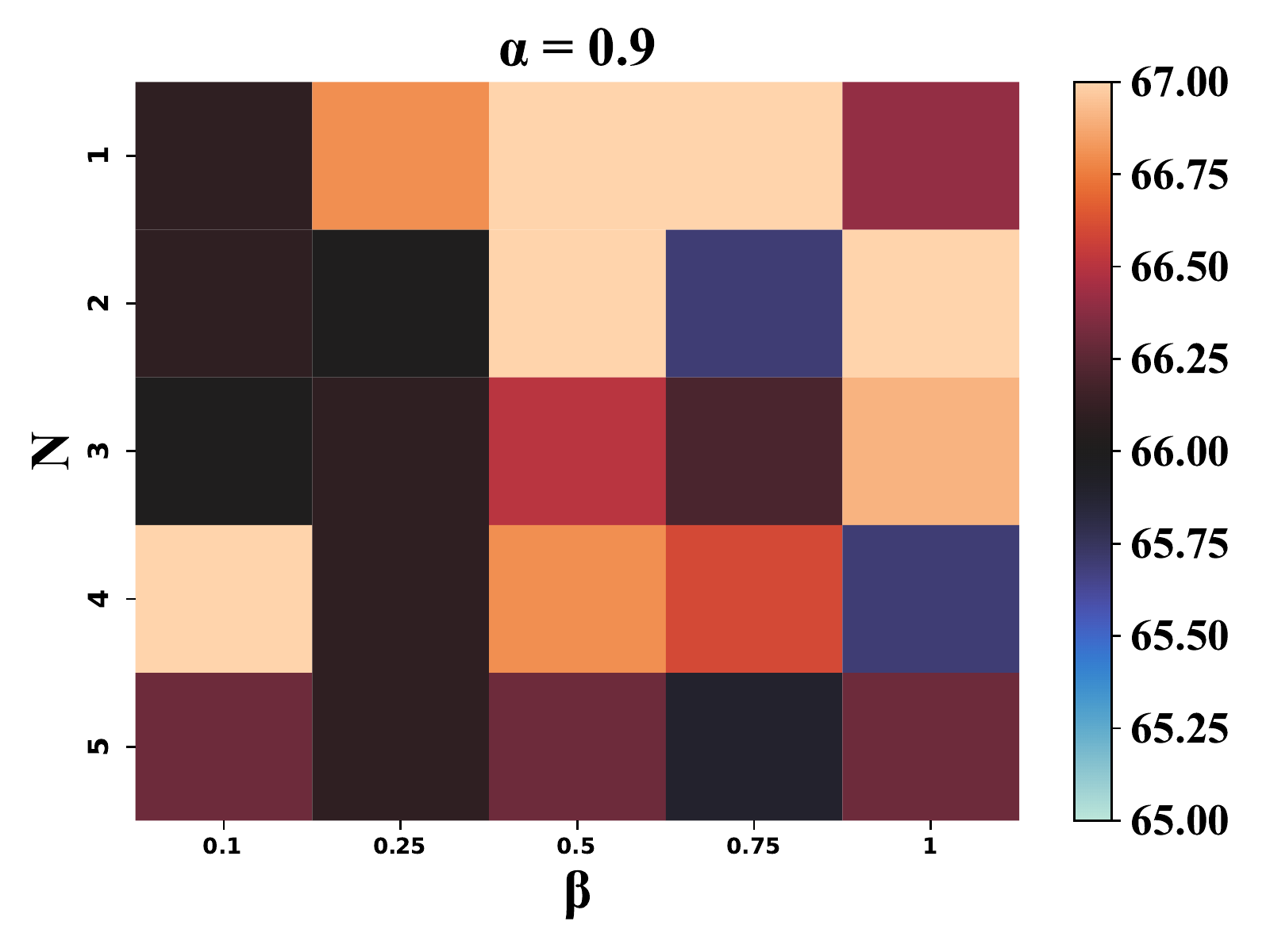}     
}    
\subfigure[]{ 
\label{fig:c}     
\includegraphics[width=0.3\textwidth]{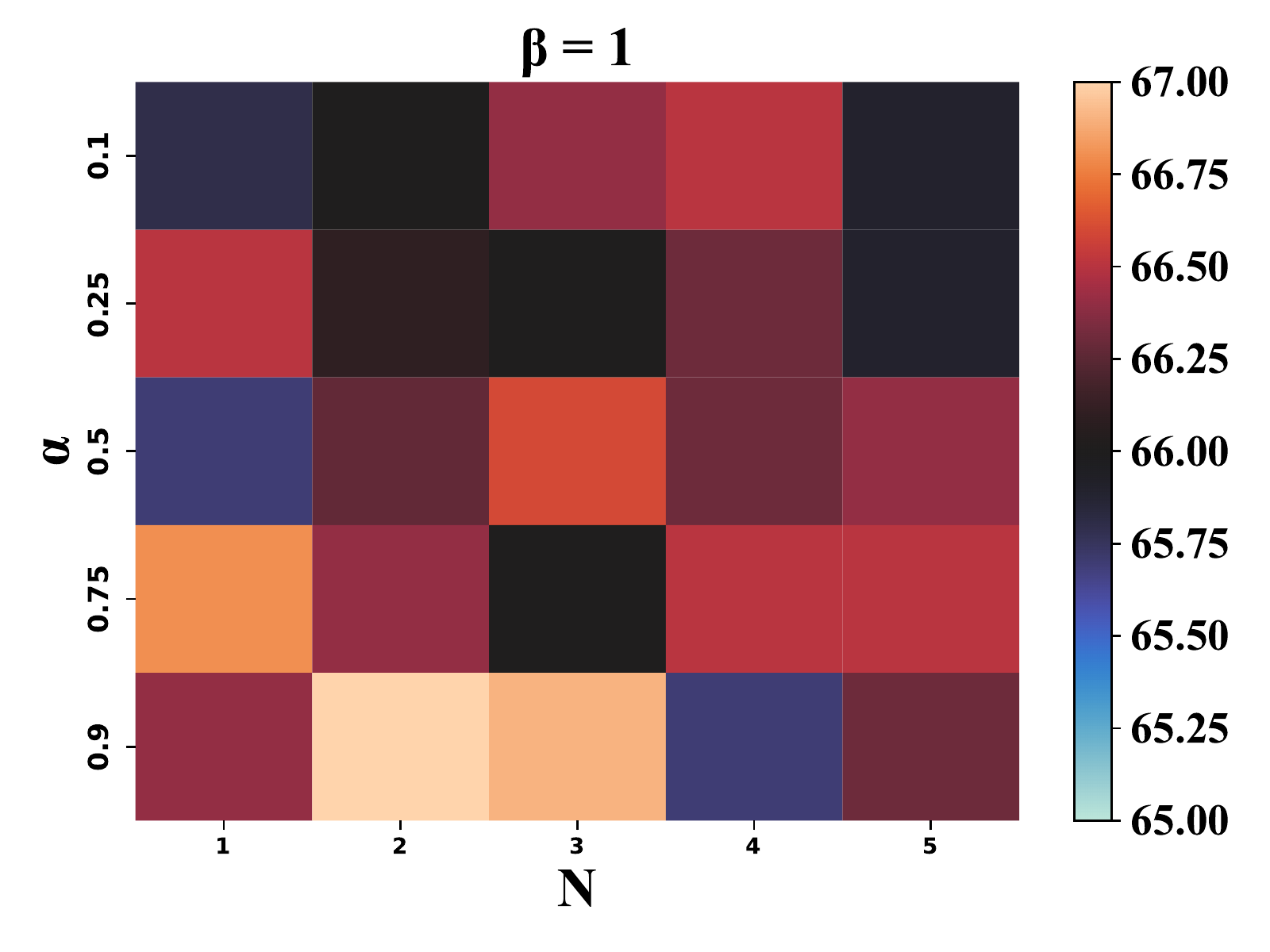}     
}   
\vskip -0.15in
\caption{F1 scores with different hyper-parameter settings. } 
\label{heatmap}     
\end{figure*}
We perform several experiments on the CPR dataset to study the influence of the hyper-parameters in our proposed CP-GCN model, and the results are shown in Figure \ref{heatmap}. The hyper-parameter $\alpha$ balances semantic and syntactic information in the dependency forest. The hyper-parameter $\beta$ balances the impact of the task-specific causal pruning module. The hyper-parameter $N$ represents the richness of semantic information.  As $(a)$, $(b)$, and $(c)$ are shown in Figure \ref{heatmap}, our proposed CP-GCN model achieves comparable performance in most settings, which indicates the robustness of our model. Specifically, CP-GCN achieves the highest F1 score 67.3 with $N=2, \alpha=0.9$, and $\beta=1$. As shown in Figure \ref{fig:c}, when $N$ decreases to 1, i.e. the semantic information decreases, CP-GCN performs best when the weight of the dependency tree, $\alpha$, decreases as well, suggesting that our model is able to balance the syntactic and semantic information. As shown in Figure \ref{fig:a}, when the weight of dependency tree $\alpha$ increases, CP-GCN performs best when the weight of task-specific causal pruning module $\beta$ increases as well. This demonstrates that there is indeed some noise in the dependency tree and our proposed task-specific causal pruning module can remove task-irrelevant information.

\end{document}